\documentclass[conference]{IEEEtran}
\IEEEoverridecommandlockouts
\usepackage{balance}
\usepackage{cite}
\usepackage{amsmath,amssymb,amsfonts}
\usepackage{algorithmic}
\usepackage{graphicx}
\usepackage{textcomp}
\usepackage{xcolor}
\usepackage[normalem]{ulem}
\useunder{\uline}{\ul}{}
\newtheorem{problem}{Problem}
\newtheorem{definition}{Definition}
\usepackage{url} 
\usepackage{booktabs} 
\usepackage{multirow} 
\def\BibTeX{{\rm B\kern-.05em{\sc i\kern-.025em b}\kern-.08em
    T\kern-.1667em\lower.7ex\hbox{E}\kern-.125emX}}
\begin{document}

\title{
Towards Global Counterfactual-Based Self-Explainable Graph Neural Networks
}

\author{\IEEEauthorblockN{Wendy Zheng\textsuperscript{\textsection}} 
\IEEEauthorblockA{
\textit{University of Virginia}\\
Charlottesville, USA \\
ncd9cf@virginia.edu}
\and
\IEEEauthorblockN{Yinhan He\textsuperscript{\textsection}}
\IEEEauthorblockA{
\textit{University of Virginia}\\
Charlottesville, USA \\
nee7ne@virginia.edu}
\and
\IEEEauthorblockN{Jundong Li}
\IEEEauthorblockA{
\textit{University of Virginia}\\
Charlottesville, USA \\
jundong@virginia.edu}
}

\maketitle

\begingroup
\renewcommand\thefootnote{\textsection}
\footnotetext{Equal contribution.}
\endgroup

\begin{abstract}
Graph Counterfactual Explanations (GCEs) have emerged as an effective approach for interpreting Graph Neural Network (GNN) predictions by identifying minimal graph modifications needed to change predictions. However, existing methods suffer from two fundamental limitations: (1) They rely on a post-hoc explanation paradigm, where a separate explainer is trained after the GNN. This sequential optimization prevents the explainer from leveraging how the GNN learned its representations, often resulting in lower-quality explanations. (2) They are largely local, generating counterfactual explanations only for individual samples. Consequently, they fail to capture generalizable patterns across the data distribution, a crucial gap for applications like drug discovery, where global principles enable systematic decision-making across thousands of cases. To address these limitations, we propose GCSGNN (\textbf{G}lobal \textbf{C}ounterfactual-based \textbf{S}elf-explainable \textbf{GNN}), a unified framework that integrates counterfactual generation into the GNN architecture, encouraging explanations that faithfully reflect the model's decision-making process. Beyond local explanations, GCSGNN discovers shared latent patterns influencing predictions across graphs, yielding global (i.e., model-level) counterfactual explanations. We further introduce an efficient latent-space counterfactual mechanism that avoids the combinatorial complexity of discrete graph manipulation, enabling scalable discovery of generalizable modification patterns. Extensive experiments on real-world datasets demonstrate that GCSGNN outperforms existing methods in generating effective counterfactual explanations. Our code is available at \url{https://github.com/wendyzheng22/gcsgnn}.
\end{abstract}


\section{Introduction}\label{sec:introduction}
\begin{figure}
  \includegraphics[width=0.3\textwidth]{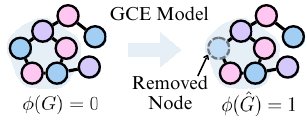}\centering
  \vspace{-10pt}
  \caption{The GCE model removes a node, thereby breaking the pentagon motif (highlighted in blue) and changing the GNN prediction to class ``1''.}\label{fig:example}
  \vspace{-10pt}
\end{figure}

Graph Counterfactual Explanations (GCEs) have demonstrated remarkable success in interpreting Graph Neural Network (GNN) predictions across a wide range of real-world applications, including brain network analysis~\cite{zannotti2026counterfactual,abrate2021counterfactual}, drug discovery~\cite{lavecchia2025explainable,wellawatte2022model, numeroso2021meg}, and credit risk prediction~\cite{crupi2024counterfactual,du2025identifying}. 
Unlike factual explanations, which identify the features responsible for a prediction, GCEs identify the minimal modifications to an input graph that change the predicted output, typically from an undesired class to a desired one. For instance, Fig.~\ref{fig:example} illustrates a GNN $\phi$ that classifies graphs containing a pentagon motif as undesired (class $0$). Removing a single node breaks the motif and changes the prediction to the desired class (class $1$), revealing the motif's influence on the model's decision.

Despite their promise, existing GCE methods face two key limitations: (1) \textbf{Post-hoc Explainer Paradigm.} Most methods employ separate explainers to generate GCEs for a trained GNN~\cite{lucic2022cfgnnexplainer,ma2022clear,huang2023global,he2024global}. Without direct access to the model's internal representations during training, these explainers may produce unstable or inconsistent explanations that poorly reflect the model's decision-making process~\cite{zhao2023towards, agarwal2023evaluating}. (2) \textbf{Limited scope}. Current methods primarily generate instance-specific (i.e., local) explanations. Although useful for understanding specific cases, such explanations fail to capture the broad patterns across the data distribution. For example, in drug discovery, local explanations may suggest modifying individual atoms or bonds but cannot readily reveal global principles, such as removing a functional group to reduce toxicity. Such global insights are crucial for scientific discovery and systematic decision-making, avoiding costly case-by-case analyses.

To address these limitations, we propose GCSGNN (\textbf{G}lobal \textbf{C}ounterfactual-based \textbf{S}elf-explainable \textbf{GNNs}), a unified framework that integrates counterfactual generation into GNN training. By jointly optimizing prediction and explanation, GCSGNN encourages faithful counterfactual explanations. Moreover, it identifies shared latent patterns across graphs to generate global (i.e., model-level) counterfactuals that capture generalizable decision-making principles. To ensure scalability, GCSGNN performs counterfactual generation in continuous latent space, avoiding the combinatorial complexity of discrete graph modifications. Our main contributions are: (1) \textbf{Latent-space Counterfactual Generation.} We develop a mechanism for generating counterfactuals in latent space, avoiding costly discrete graph manipulation. (2) \textbf{Global Self-explainable Framework.} We introduce GCSGNN, which jointly learns predictions and counterfactual explanations that uncover generalizable patterns. (3) \textbf{Extensive evaluation.} Experiments on real-world datasets demonstrate that GCSGNN generates valid, transferable global counterfactuals.

\section{Preliminaries and Problem Definition}\label{sec: preliminaries}
In this section, we first introduce the notation and define \emph{counterfactual graph embedding edits (CGEEs)}, then formalize  global self-explainable graph counterfactual explanation.

Let $\mathcal{G}$ denote a graph dataset, where each graph $G=(\mathbf{X},\mathbf{A},\mathbf{E})$ contains $n$ nodes, with node features $\mathbf{X}\in\mathbb{R}^{n\times d_X}$, adjacency matrix $\mathbf{A}\in\mathbb{R}^{n\times n}$, and edge features $\mathbf{E}\in\mathbb{R}^{n\times n\times d_E}$. All graphs are padded to a fixed maximum node size. A GNN $\phi=f_p\circ f_e$ consists of an encoder $f_e$ that maps $G$ to a graph embedding $\mathbf{h}_G\in\mathbb{R}^{d}$ and a predictor $f_p$ that outputs class $0$ (undesired) or $1$ (desired).

As introduced in Section~\ref{sec:introduction}, our goal is to discover global counterfactual patterns that explain a GNN’s behavior across multiple graphs. While prior methods search for structural modifications directly in graph space~\cite{huang2023global,he2024global}, such discrete optimization introduces combinatorial complexity. Instead, we propose operating in continuous latent space by identifying influential embedding channels and replacing their values with learned counterfactual sub-embeddings.

\begin{figure}
  \includegraphics[width=0.35\textwidth]{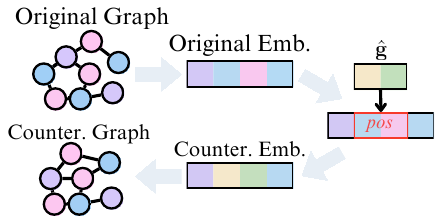}\centering
  \vspace{-3pt}
  \caption{The CGEE replaces the sub-embedding located at the significant channels ``pos'' in the graph embedding with the counterfactual $\hat{\mathbf{g}}$.}
  \label{fig:cgee}
  \vspace{-10pt}
\end{figure}

Formally, as shown in Fig.~\ref{fig:cgee}, a counterfactual graph embedding edit (CGEE) is a tuple $s=(pos,\hat{\mathbf{g}})$, where $pos\subseteq\{1,\ldots,d\}$ denotes the positions (i.e., indices) of the significant embedding channels and $\hat{\mathbf{g}}\in\mathbb{R}^{d_s}$ is a counterfactual sub-embedding, with $d_s=|pos|\ll d$. Given $\mathbf{h}_G=f_e(G)$, applying $s$ replaces the values at $pos$ with $\hat{\mathbf{g}}$ to obtain a counterfactual embedding $\hat{\mathbf{h}}_G^s$, which is decoded into a counterfactual graph $\hat{G}_s$. A set of CGEEs $\mathcal{S}$ constitutes a global explanation, while the decoded counterfactuals for individual graphs provide local explanations. We evaluate the effectiveness of a CGEE set through its \emph{coverage}.

\begin{definition}\label{def:coverage}
\textbf{Coverage of a CGEE Set.} Given a GNN $\phi$, a set of undesired graphs $\mathcal{G}_0=\{G\in\mathcal{G}\mid\phi(G)=0\}$, and a CGEE set $\mathcal{S}$, coverage measures the proportion of graphs in $\mathcal{G}_0$ for which applying one CGEE $s \in \mathcal{S}$ yields a ``valid'' counterfactual (i.e., the generated counterfactual graph is classified by the GNN as desired): 
\begin{equation}
 \mathbf{coverage}(\mathcal{S})=\frac{|\{G\in\mathcal{G}_0 \mid \exists s \in \mathcal{S}, \phi(\hat{G}_s) = 1 \}|}{|\mathcal{G}_0|}
\end{equation}
\end{definition} 

By replacing discrete graph edits with continuous latent-space optimization, CGEEs transform global counterfactual discovery into a continuous optimization problem, allowing counterfactual generation to be integrated with GNN training. We formalize this objective as follows. 

\begin{problem}\label{pro:self_global_gnn}
\textbf{Global Self-Explainable Graph Counterfactual Explanation:}
Given a graph dataset $\mathcal{G}$ with ground-truth labels $y_G \in Y$ and a budget $k \ll |\mathcal{G}|$, learn a self-explainable GNN $\phi_{\mathcal{S}} : \mathcal{G} \rightarrow \{0,1\}$ such that:
\begin{enumerate}
\item $\phi_{\mathcal{S}}$ accurately predicts the labels $y_G$ for graphs in $\mathcal{G}$
\item the associated CGEE set $\mathcal{S}$ maximizes coverage under the budget constraint $|\mathcal{S}| = k$
\end{enumerate}
Formally,
\begin{gather*}
\max_{\phi_{\mathcal{S}^*}} |\{G \in \mathcal{G} \mid \phi_{\mathcal{S}^*}(G) = y_G\}|, \\
\quad \text{s.t.} \quad
\mathcal{S}^* = \arg\max_{\mathcal{S}} \mathbf{coverage}(\mathcal{S}) \text{ and } |\mathcal{S}| = k.
\end{gather*}
\end{problem}

\section{Methodology}
\begin{figure*}
 \centering
 \includegraphics[width=0.77\textwidth]{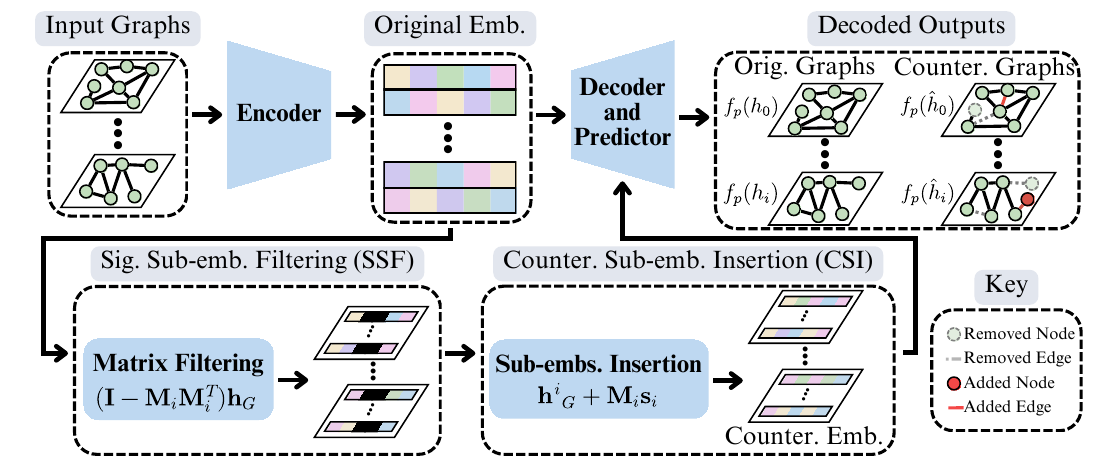}
 \vspace{-5pt}
 \caption{An overview of the proposed GCSGNN model.}
 \vspace{-10pt}
 \label{fig:overview}
\end{figure*}
In this section, we present the GCSGNN architecture, followed by its counterfactual generation mechanism, graph reconstruction process, and training objectives.

\subsection{Model Overview}
As shown in Fig.~\ref{fig:overview}, GCSGNN consists of four components: (1) a \textbf{Graph Encoder} ($f_e$) that maps input graphs to latent embeddings; (2) a \textbf{CGEE Generator} ($f_c$) that produces counterfactual embeddings by replacing significant embedding channels; (3) a \textbf{Graph Decoder} ($f_d$) that reconstructs graphs from original and counterfactual embeddings; and (4) a \textbf{Graph Predictor} ($f_p$) that classifies these embeddings. These components are jointly trained to enable self-explainable GNN predictions and global counterfactual explanations.

\subsection{CGEE Generation}\label{subsec:global_cf_gen}
As defined in Section~\ref{sec: preliminaries}, global counterfactual explanations are represented by a set of $k$ Counterfactual Graph Embedding Edits (CGEEs) $\{(pos_i,\hat{\mathbf{g}}_i)\}_{i=1}^k$. We learn these CGEEs using two parameters: a mask tensor $\mathbf{M}\in\mathbb{R}^{d\times d_s\times k}$ that selects significant embedding channels and a matrix $\mathbf{S}\in\mathbb{R}^{k\times d_s}$ containing counterfactual sub-embeddings, where $d_s\ll d$.

Each mask $\mathbf{M}_i=\mathbf{M}_{:,:,i}\in\mathbb{R}^{d\times d_s}$ consists of one-hot columns identifying $d_s$ significant channels of the $i$th CGEE. Its corresponding counterfactual sub-embedding is $\mathbf{s}_i=\mathbf{S}_{i,:}^T\in\mathbb{R}^{d_s}$. Given a graph embedding $\mathbf{h}_G=f_e(G)$, we generate its $i$th counterfactual by replacing the selected channels: 
\begin{equation}
\hat{\mathbf{h}}_G^i =
(\mathbf{I}-\mathbf{M}_i\mathbf{M}_i^T)\mathbf{h}_G
+\mathbf{M}_i\mathbf{s}_i,
\end{equation}
where $\mathbf{I}\in\mathbb{R}^{d\times d}$ is the identity matrix. The first term removes the original values at the selected channels, while the second inserts the counterfactual sub-embedding.

To encourage $\hat{\mathbf{h}}_G^i$ to be classified as desired (class $1$), we minimize the counterfactual prediction loss:

\begin{equation}\label{eqn:cf_pred_loss}
\mathcal{L}_c = -\frac{1}{k|\mathcal{G}_0|}\sum_{i\in\{1,...,k\},G\in\mathcal{G}_0}\log{P(f_p(\hat{\mathbf{h}}_G^i) = 1)}.
\end{equation}
By averaging the counterfactual prediction loss across all CGEEs, we provide uniform supervision to learn GCEs without favoring particular edits during optimization. We further encourage the counterfactual sub-embeddings to capture characteristics of the desired class:

\begin{equation}\label{eqn:embedding_loss}
 \mathcal{L}_e = \frac{1}{| \mathcal{G}_1|}\sum_{G \in \mathcal{G}_1}\min_i |\mathbf{h}_G-\mathbf{v}_i|_2^2,
\end{equation}
where $\mathcal{G}_1$ denotes the set of desired graphs and $\mathbf{v}_i=\mathbf{M}_i\mathbf{s}_i\in\mathbb{R}^d$ is the expanded counterfactual sub-embedding. This loss encourages the learned CGEEs to capture shared embedding patterns of the desired class.

\subsection{Counterfactual Graph Reconstruction}
The graph decoder $f_d$ reconstructs graph structures from original and counterfactual embeddings. Given $\mathbf{h}_G$ and the counterfactual embedding set
\begin{equation}
\hat{\mathbf{H}}_G =
[\hat{\mathbf{h}}_G^1,\ldots,\hat{\mathbf{h}}_G^k]
\in\mathbb{R}^{d\times k},
\end{equation}
the decoder reconstructs the original graph $G=(\mathbf{X},\mathbf{A},\mathbf{E})$ and counterfactual graphs $\{\hat{G}_i= (\hat{\mathbf{X}}_i, \hat{\mathbf{A}}_i, \hat{\mathbf{E}}_i)\}_{i=1}^{k}$.

Specifically, an MLP layer expands each counterfactual embedding $\hat{\mathbf{h}}_G^i$ into a combined node–edge representation $\hat{\mathbf{C}}_i\in\mathbb{R}^{n\times d}$. A second MLP extracts node embeddings $\hat{\mathbf{N}}_i\in\mathbb{R}^{n\times d}$, while aggregated edge embeddings are obtained as $\hat{\mathbf{E}}_{agg,i}=\hat{\mathbf{C}}_i-\hat{\mathbf{N}}_i$. Separate MLPs then reconstruct node features $\hat{\mathbf{X}}_i$ and edge features $\hat{\mathbf{E}}_i$. The adjacency matrix $\hat{\mathbf{A}}_i$ is obtained by thresholding the decoded edge features, with nonzero feature vectors indicating edges.

The decoder is trained using the reconstruction loss
\begin{equation}
\mathcal{L}r =
\mathcal{L}_{r,A}+\mathcal{L}_{r,E}+\mathcal{L}_{r,X},
\end{equation}
where $\mathcal{L}_{r,A}$ uses weighted binary cross-entropy for adjacency reconstruction, $\mathcal{L}_{r,E}$ uses multi-class cross-entropy for edge features, and $\mathcal{L}_{r,X}$ uses weighted multi-class cross-entropy for discrete node features or the $L_2$ norm for continuous features.


\subsection{Model Objective Function}
Following Problem~\ref{pro:self_global_gnn}, GCSGNN also optimizes graph classification, so the graph prediction loss is

\begin{equation}\label{eqn:graph_pred_loss}
\mathcal{L}_g = -\frac{1}{|\mathcal{G}|}\sum_{G\in\mathcal{G}}\log{P(f_p(\mathbf{h}_G) = y_G)}.
\end{equation}

We additionally introduce two regularizers for the CGEE parameters. First, to encourage each column of $\mathbf{M}_i$ to be one-hot, we define the mask regularization loss

\begin{equation}\label{eqn:mask_loss}
\mathcal{L}_m = \Sigma_{i=1}^{k}\Sigma_{l=1}^{d_s}\Bigg[\left|\mathbf{1} -\max_p{\mathbf{M}_{p,l,i}} \right| +\left|\mathbf{1} - \Sigma_{p=1}^{d}{|\mathbf{M}_{p,l,i}|} \right|\Bigg].
\end{equation}

The first term encourages a maximum value of one, while the second encourages each column's absolute sum to equal one, promoting valid channel selections. Second, to encourage CGEEs that capture distinct counterfactual patterns, we define

\begin{equation}\label{eqn}
\mathcal{L}_i =
\frac{1}{kd_s}
\left|
\mathbf{V}\mathbf{V}^T-\mathbf{D}
\right|_1, \mathbf{V}=[\mathbf{v}_1,\ldots,\mathbf{v}_k]^T\in\mathbb{R}^{k\times d}
\end{equation}
where $\mathbf{V}$ is the set of expanded counterfactual sub-embeddings and $\mathbf{D}$ is the diagonal matrix with $\mathbf{D}_{ii}=|\mathbf{v}_i|_2^2$. 

Combining all defined losses gives
\begin{equation}
 \mathcal{L} = \alpha\mathcal{L}_g + \beta\mathcal{L}_c +\gamma \mathcal{L}_r+\delta \mathcal{L}_e + \mathcal{L}_m + \mathcal{L}_i.
\end{equation}
where $\alpha$, $\beta$, $\gamma$, and $\delta$ control each objective's contribution. During training, $\mathcal{L}_c$ is computed directly on edited graph embeddings. However, during evaluation, each counterfactual embedding is decoded into a graph, which is subsequently re-encoded and classified by the complete GNN. Thus, counterfactual validity and coverage are measured on reconstructed graphs rather than directly on edited embeddings.

\section{Experiments}

\begin{table*}[]
\caption{Performance of GCSGNN compared to baselines, averaged over 5 runs. Bold indicates best results; underlined marks runner-ups. ``OOT'' denotes experiments exceeding 72 hours.}\label{tab:main_table}
\centering
\resizebox{\textwidth}{!}{
\begin{tabular}{ccccccccc}
\toprule
                         &                    & CF-GNNExp.   & GCFExp.           & GNNExp.               & InduCE               & ProtGNN       & RCExp.               & GCSGNN                 \\ \midrule
\multirow{2}{*}{AIDS}    & Cov. $~\uparrow$   & 0.00 ± 0.00  & {\ul 2.82 ± 0.93} & 0.00 ± 0.00           & 0.00 ± 0.00          & 0.00 ± 0.00   & 0.04 ± 0.03          & \textbf{80.28 ± 12.65} \\
                         & Prox.$~\downarrow$ & N/A          & {\ul 8.10 ± 0.23} & N/A                   & N/A                  & N/A           & \textbf{4.69 ± 1.80} & 16.31 ± 1.56           \\ \midrule
\multirow{2}{*}{BZR\_MD} & Cov. $~\uparrow$   & 25.38 ± 6.47 & 10.64 ± 2.91      & {\ul 67.05 ± 3.28}    & 16.54 ± 11.08        & 0.00 ± 0.00   & 59.10 ± 4.89         & \textbf{93.72 ± 6.50}  \\
                         & Prox.$~\downarrow$ & 12.18 ± 0.29 & {\ul 5.92 ± 0.57} & 19.10 ± 0.06          & \textbf{4.52 ± 0.91} & N/A           & 9.75 ± 0.24          & 15.14 ± 0.68           \\ \midrule
\multirow{2}{*}{MUTAG}   & Cov. $~\uparrow$   & 0.48 ± 0.65  & 34.52 ± 6.81      & {\ul 87.58 ± 2.58}    & 0.65 ± 0.79          & 0.00 ± 0.00   & 26.61 ± 3.71         & \textbf{99.84 ± 0.32}  \\
                         & Prox.$~\downarrow$ & 10.25 ± 0.18 & 8.79 ± 0.37       & 17.98 ± 0.05          & \textbf{5.62 ± 1.18} & N/A           & {\ul 8.65 ± 0.20}    & 14.15 ± 0.49           \\ \midrule
\multirow{2}{*}{CIFAR10} & Cov. $~\uparrow$   & 0.18 ± 0.12  & OOT               & {\ul 63.38 ± 8.73}    & 0.00 ± 0.00          & 12.64 ± 11.07 & 6.78 ± 2.50          & \textbf{85.92 ± 18.40} \\
                         & Prox.$~\downarrow$ & 66.35 ± 0.25 & OOT               & 110.44 ± 0.76         & N/A                  & 116.12 ± 1.14 & {\ul 56.31 ± 0.04}   & \textbf{21.45 ± 13.13} \\ \midrule
\multirow{2}{*}{MNIST}   & Cov. $~\uparrow$   & 1.99 ± 0.27  & OOT               & \textbf{98.49 ± 0.10} & 0.12 ± 0.09          & 0.00 ± 0.00   & 3.93 ± 0.72          & {\ul 97.80 ± 2.72}     \\
                         & Prox.$~\downarrow$ & 85.21 ± 0.22 & OOT               & 98.13 ± 0.00          & {\ul 78.19 ± 1.53}   & N/A           & 83.63 ± 0.23         & \textbf{42.63 ± 8.08}  \\ \bottomrule
\end{tabular}}
\vspace{-10pt}
\end{table*}


In this section, we evaluate GCSGNN through extensive experiments addressing four research questions: \textbf{RQ1.} How does GCSGNN compare to state-of-the-art methods? \textbf{RQ2.} How do hyperparameters affect performance? \textbf{RQ3.} How do its components contribute to performance? \textbf{RQ4.} Can GCSGNN provide meaningful global counterfactual insights?

\subsection{Experiment Setup}\label{sec:exp_setup}
\noindent\textit{Datasets.} We evaluate on five real-world datasets: AIDS, BZR\_MD, and MUTAG~\cite{Morris2020} contain molecular graphs with atoms as nodes and chemical bonds as edges, where the majority class is designated as undesired. CIFAR10~\cite{alex2009learning} and MNIST~\cite{deng2012mnist} are converted into graphs by treating pixels as nodes and connecting neighboring pixels. For CIFAR10, we select ``airplane'' (undesired) and ``bird'' (desired), while for MNIST, we select ``0'' (undesired) and ``1'' (desired).

\noindent\textit{Baselines.} We adopt six state-of-the-art methods as baselines: (1) CF-GNNExplainer~\cite{lucic2022cfgnnexplainer}, a local counterfactual method adapted for graph-level explanations; (2) GCFExplainer~\cite{huang2023global}, which generates global counterfactuals using vertex-reinforced random walks; (3) GNNExplainer~\cite{ying2019gnnexplainer}, a local factual method modified for counterfactual generation; (4) InduCE~\cite{verma2024induce}, a policy-based local counterfactual method adapted to graph classification; (5) ProtGNN~\cite{zhang2022protgnn}, a prototype-based self-explainable GNN whose objective we adapt for global counterfactual generation; and (6) RCExplainer~\cite{bajaj2021robust}, which approximates shared decision boundaries to generate robust counterfactuals. All post-hoc baselines use a trained GCSGNN as the underlying prediction model for a consistent comparison.

\noindent\textit{Evaluation Metrics.} We evaluate coverage, the proportion of undesired graphs for which valid counterfactuals are generated, and proximity, the squared difference between the original and counterfactual graphs' one-hot-encoded node features, edge features, and adjacency matrices. Higher coverage and lower proximity indicate better performance.

\subsection{RQ1: Effectiveness of GCSGNN}
Table~\ref{tab:main_table} compares GCSGNN with the baselines, yielding two key observations. (1) \textbf{Superior Global Performance.} GCSGNN consistently outperforms global baselines (i.e., ProtGNN, RCExplainer, and GCFExplainer) in coverage, achieving 80.28\%, 93.72\%, and 99.84\% on AIDS, BZR\_MD, and MUTAG, respectively. Its advantage extends to image datasets, achieving 85.92\% coverage on CIFAR10 compared to 12.64\% for the second-best global baseline, ProtGNN. On MNIST, GCSGNN reaches 97.80\%, outperforming RCExplainer (3.93\%). These results demonstrate GCSGNN's ability to generate valid global counterfactuals across diverse graph datasets. (2) \textbf{Coverage–Proximity Trade-off.} Proximity alone can be misleading, as methods with low coverage may achieve small edit distances. GCSGNN achieves a favorable balance between both metrics. On BZR\_MD and MUTAG, it exceeds 90\% coverage while achieving lower proximity than GNNExplainer. Similarly, on CIFAR10 and MNIST, GCSGNN obtains proximity values of 21.45 and 42.63, compared to 110.44 and 98.13 for GNNExplainer. Although GNNExplainer achieves slightly higher coverage on MNIST, the explanations from GCSGNN requires significantly less edits. 

\begin{figure}
 \centering
 \includegraphics[width=0.5\textwidth]{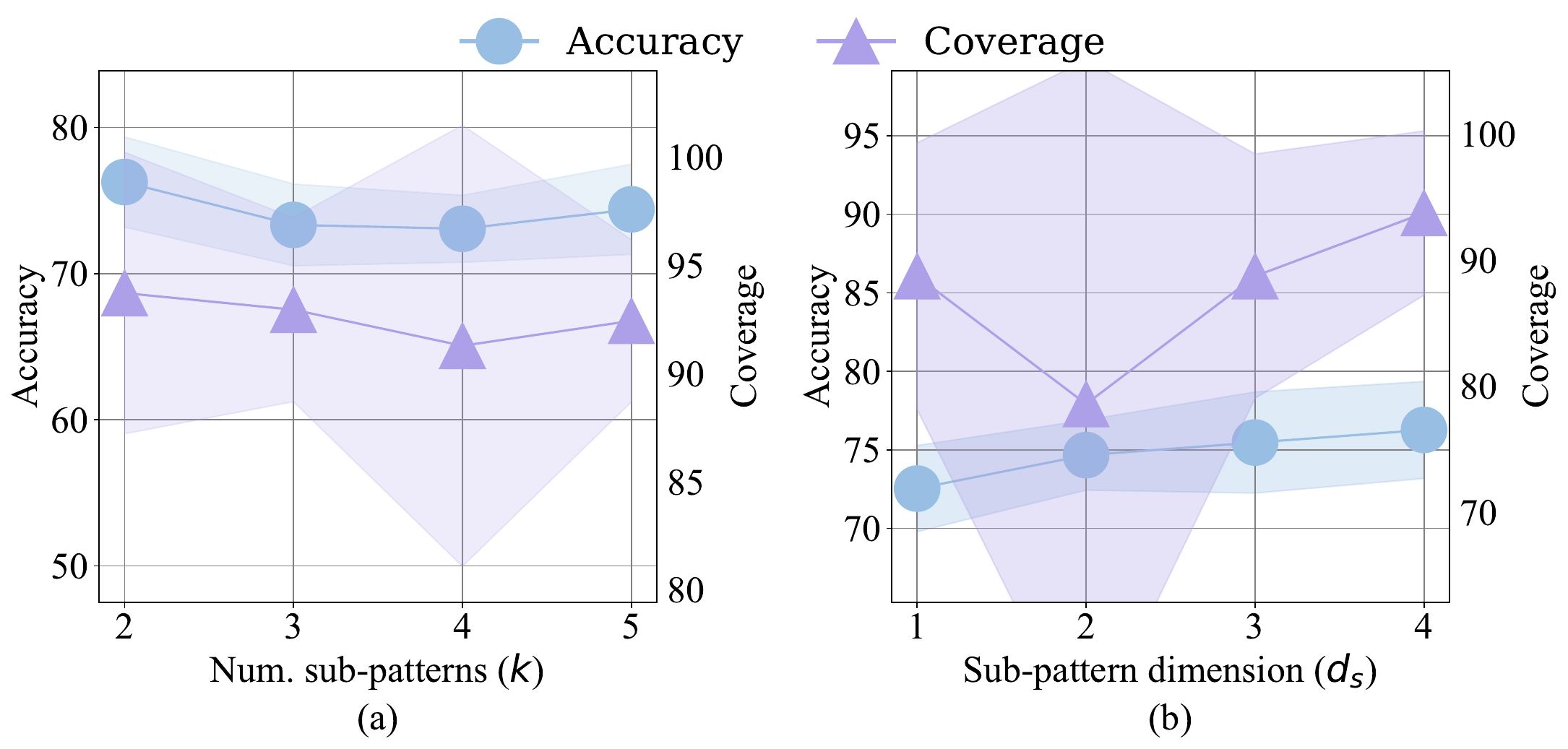}
 \vspace{-15pt}
 \caption{Coverage and accuracy on BZR\_MD for varying $k$ and $d_s$ values.}
 \vspace{-10pt}
 \label{fig:param}
\end{figure}

\begin{figure*}[]
 \centering
 \includegraphics[width=0.85\textwidth]{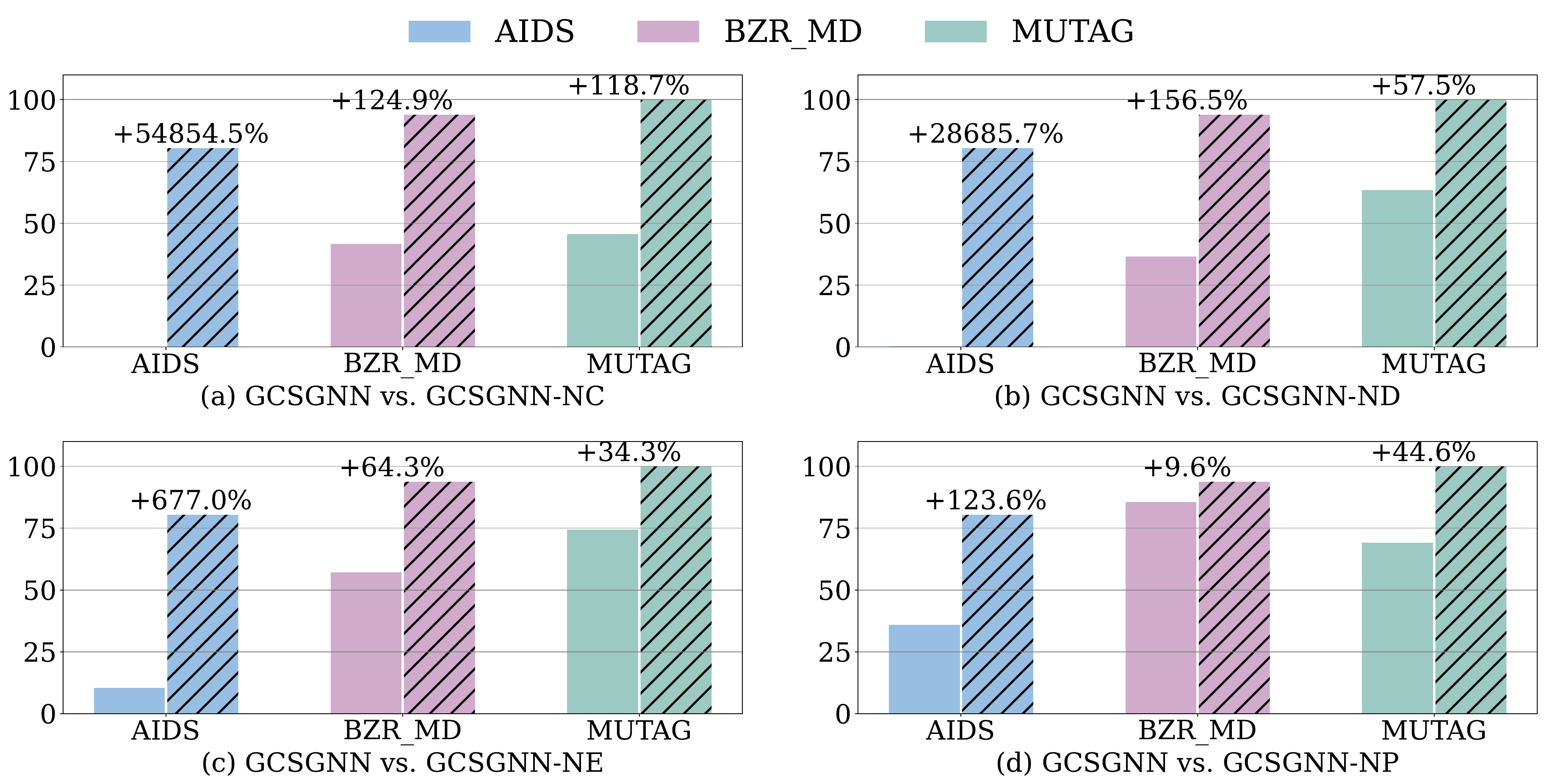}
 \vspace{-5pt}
 \caption{Ablation study. The hatched bars represent the coverage for GCSGNN.}
 \vspace{-10pt}
 \label{fig:ablation}
\end{figure*}

\subsection{RQ2: Parameter Analysis}
We analyze the sensitivity of GCSGNN to two key hyperparameters on the BZR\_MD dataset: the number of sub-embeddings $k \in \{2,3,4,5\}$ and the sub-embedding dimension $d_s \in \{1,2,3,4\}$. As shown in Fig. ~\ref{fig:param}, we make the following observations: (1) \textbf{Number of sub-embeddings $k$}: Performance peaks at $k=2$, which aligns with the dataset size and complexity. Larger values of $k$ lead to redundant sub-embeddings and degrade both coverage and accuracy. Accuracy remains relatively stable across values of $k$, indicating low sensitivity. (2) \textbf{Sub-embedding dimension $d_s$}: Increasing $d_s$ generally improves both accuracy and coverage, with both peaking at $d_s=4$. This is likely because BZR\_MD graphs are more dense (average 21.30 nodes and 225.06 edges), which may require higher-dimensional edits to induce prediction changes.

\subsection{RQ3: Ablation Study}
To understand each component's contribution, we evaluate four variants (Fig.~\ref{fig:ablation}): GCSGNN-NC (\textbf{N}o \textbf{C}ounterfactuals), GCSGNN-ND (\textbf{N}o \textbf{D}ecoder), GCSGNN-NE (\textbf{N}o \textbf{E}ncoder), and GCSGNN-NP (\textbf{N}o \textbf{P}redictor). GCSGNN-NC follows a post-hoc training paradigm, where counterfactual components are trained after the prediction model. The remaining variants use the corresponding component pretrained under the post-hoc setting, freeze its parameters, and jointly train the other components, which are randomly initialized.

Across datasets, GCSGNN-NC shows the largest coverage degradation, highlighting the importance of integrating counterfactual generation into GNN training. Freezing the pretrained encoder or decoder also substantially reduces coverage, demonstrating the importance of jointly learning graph representations and reconstructions. In contrast, freezing the predictor causes a smaller decline, suggesting that the counterfactual embeddings remain compatible with the learned prediction space. Overall, these results demonstrate that GCSGNN benefits from jointly optimizing its prediction and explanation components rather than training them separately.

\begin{figure*}
 \centering
 \includegraphics[width=0.85\textwidth]{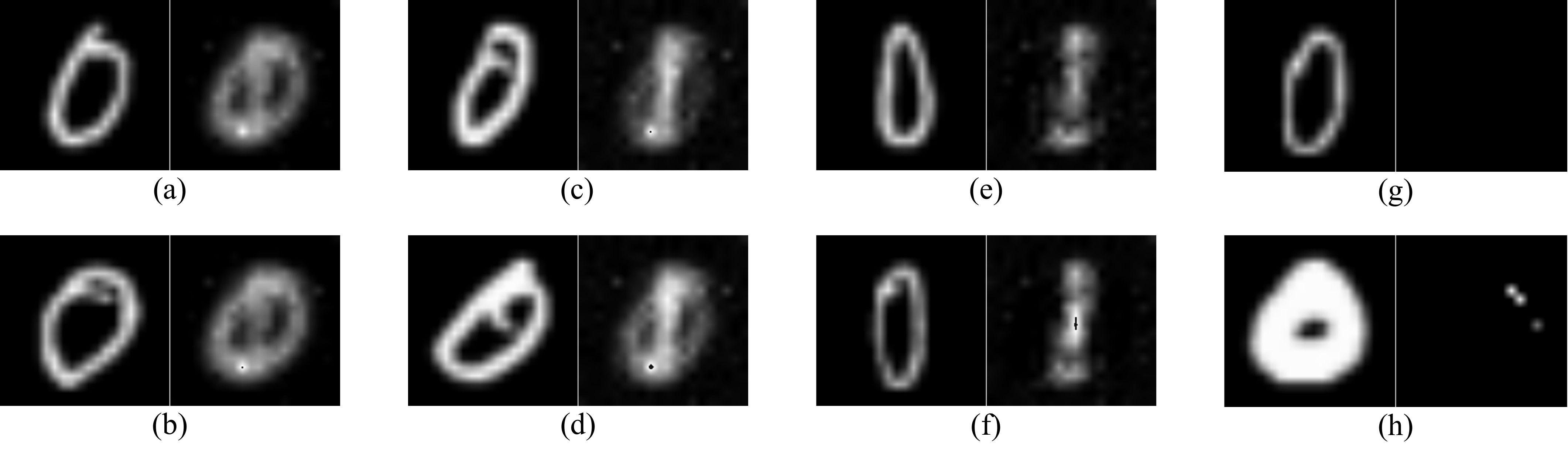}
 \vspace{-10pt}
 \caption{
 Each subfigure (except (g) and (h)) shows an MNIST sample of class ``0'' (left) and its GCSGNN-generated counterfactual predicted as class ``1'' (right). Pairs (a)–(b), (c)–(d), and (e)–(f) share the same counterfactual sub-embedding. (g) and (h) show GNNExplainer-generated counterfactuals for comparison.}
 \vspace{-10pt}
 \label{fig:case_study}
\end{figure*}

\subsection{RQ4: Case Study}
We evaluate GCSGNN's ability to generate meaningful GCEs through a case study (Fig.~\ref{fig:case_study}(a)-(f)), where digit ``0'' graphs are transformed into counterfactuals classified as digit ``1''. Negative node attributes are clipped to zero for visualization. Each pair of examples, (a)–(b), (c)–(d), and (e)–(f), shares the same counterfactual sub-embedding.

Rather than applying a universal transformation, GCSGNN learns three distinct counterfactual patterns adapted to different input structures. For wide, thick zeros (a)–(b), it introduces a central vertical stroke while weakening the surrounding loop. For medium-width zeros (c)–(d), it more aggressively suppresses the circular structure, leaving a dominant vertical stroke with faint residual traces. For narrow zeros (e)–(f), whose geometry already resembles a vertical shape, it nearly eliminates the loop, producing a clean digit ``1''. These results demonstrate that GCSGNN captures structural similarities across graphs and learns reusable counterfactual patterns that induce consistent modifications within each group.

We further compare GCSGNN with GNNExplainer in Fig.~\ref{fig:case_study}(g)–(h). Unlike GCSGNN's coherent transformations, GNNExplainer produces scattered pixels that successfully flip predictions but do not resemble the target digit nor the original image. Such perturbations suggest that local optimization may exploit the classifier's sensitivity to small, semantically meaningless changes. In contrast, GCSGNN learns global modification patterns that preserve semantic coherence across similar inputs, providing more interpretable insights into how graph structures influence model predictions.

\section{Related Work}
\subsection{Post-hoc Counterfactual-based Explanations}
Post-hoc counterfactual methods identify graph modifications that change predictions of pretrained GNNs. Local methods generate instance-specific counterfactuals through graph perturbations or generative models~\cite{lucic2022cfgnnexplainer,ma2022clear,verma2024induce}, while global methods identify representative counterfactuals or shared modification rules across graphs~\cite{bajaj2021robust,huang2023global,he2024global}. However, this post-hoc paradigm limits the explainer's access to the model's internal decision-making process. In contrast, GCSGNN jointly optimizes graph prediction and global counterfactual generation in continuous latent space.

\subsection{Self-explainable GNNs}
Self-explainable GNNs integrate explanation generation into model training, typically identifying influential subgraphs or prototypes ~\cite{zhang2022protgnn, dai2021selfexplainable, deng2024self}. Recent methods extend this paradigm through class-level subgraph explanations~\cite{liu2026towards} and interpretable logical rules~\cite{ragno2026logic}. GCSGNN extends self-explainability to global counterfactual reasoning by learning reusable latent-space edits to change model prediction.

\balance

\section{Conclusion}


In this paper, we propose GCSGNN, a novel framework for global counterfactual-based self-explainable graph neural networks. By integrating counterfactual generation into GNN training, GCSGNN jointly optimizes prediction and explanation, addressing the limitations of post-hoc approaches. Through latent-space counterfactual edits, GCSGNN efficiently captures shared modification patterns across graphs. Extensive experiments demonstrate its effectiveness in generating valid, transferable global counterfactuals. Overall, GCSGNN provides an effective approach to understanding GNN decision-making through reusable counterfactual explanations.

\section*{Acknowledgment}
This work is supported in part by the National Science Foundation (NSF) under Grants 2144209, 2223769, 2228534, 2411248, and 2601942; the Office of Naval Research (ONR) under Grant N000142412636; the Commonwealth Cyber Initiative (CCI) under Grant HV-4Q26-073.

\bibliographystyle{IEEEtran}
\bibliography{example_references}

@article{deng2012mnist,
  title={The mnist database of handwritten digit images for machine learning research [best of the web]},
  author={Deng, Li},
  journal={IEEE signal processing magazine},
  volume={29},
  number={6},
  pages={141--142},
  year={2012},
  publisher={IEEE}
}

@inproceedings{zhang2022protgnn,
  title={Protgnn: Towards self-explaining graph neural networks},
  author={Zhang, Zaixi and Liu, Qi and Wang, Hao and Lu, Chengqiang and Lee, Cheekong},
  booktitle={Proceedings of the AAAI Conference on Artificial Intelligence},
  volume={36},
  number={8},
  pages={9127--9135},
  year={2022}
}

@inproceedings{lucic2022cfgnnexplainer,
  title={Cf-gnnexplainer: Counterfactual explanations for graph neural networks},
  author={Lucic, Ana and Ter Hoeve, Maartje A and Tolomei, Gabriele and De Rijke, Maarten and Silvestri, Fabrizio},
  booktitle={International Conference on Artificial Intelligence and Statistics},
  pages={4499--4511},
  year={2022},
  organization={PMLR}
}

@article{ying2019gnnexplainer,
  title={Gnnexplainer: Generating explanations for graph neural networks},
  author={Ying, Zhitao and Bourgeois, Dylan and You, Jiaxuan and Zitnik, Marinka and Leskovec, Jure},
  journal={Advances in neural information processing systems},
  volume={32},
  year={2019}
}

@inproceedings{dai2021selfexplainable,
  title={Towards self-explainable graph neural network},
  author={Dai, Enyan and Wang, Suhang},
  booktitle={Proceedings of the 30th ACM International Conference on Information \& Knowledge Management},
  pages={302--311},
  year={2021}
}

@inproceedings{Morris2020,
    title={TUDataset: A collection of benchmark datasets for learning with graphs},
    author={Christopher Morris and Nils M. Kriege and Franka Bause and Kristian Kersting and Petra Mutzel and Marion Neumann},
    booktitle={ICML 2020 Workshop on Graph Representation Learning and Beyond (GRL+ 2020)},
    archivePrefix={arXiv},
    eprint={2007.08663},
    url={www.graphlearning.io},
    year={2020}
}

@inproceedings{deng2024self,
  title={Self-Interpretable Graph Learning with Sufficient and Necessary Explanations},
  author={Deng, Jiale and Shen, Yanyan},
  booktitle={Proceedings of the AAAI Conference on Artificial Intelligence},
  volume={38},
  number={10},
  pages={11749--11756},
  year={2024}
}

@inproceedings{zhao2023towards,
  title={Towards faithful and consistent explanations for graph neural networks},
  author={Zhao, Tianxiang and Luo, Dongsheng and Zhang, Xiang and Wang, Suhang},
  booktitle={Proceedings of the Sixteenth ACM International Conference on Web Search and Data Mining},
  pages={634--642},
  year={2023}
}

@article{agarwal2023evaluating,
  title={Evaluating explainability for graph neural networks},
  author={Agarwal, Chirag and Queen, Owen and Lakkaraju, Himabindu and Zitnik, Marinka},
  journal={Scientific Data},
  volume={10},
  number={1},
  pages={144},
  year={2023},
  publisher={Nature Publishing Group UK London}
}

@inproceedings{huang2023global,
  title={Global counterfactual explainer for graph neural networks},
  author={Huang, Zexi and Kosan, Mert and Medya, Sourav and Ranu, Sayan and Singh, Ambuj},
  booktitle={Proceedings of the Sixteenth ACM International Conference on Web Search and Data Mining},
  pages={141--149},
  year={2023}
}

@article{alex2009learning,
  title={Learning multiple layers of features from tiny images},
  author={Krizhevsky, Alex},
  year={2009},
  publisher={Toronto, ON, Canada}
}

@article{ma2022clear,
  title={Clear: Generative counterfactual explanations on graphs},
  author={Ma, Jing and Guo, Ruocheng and Mishra, Saumitra and Zhang, Aidong and Li, Jundong},
  journal={Advances in neural information processing systems},
  volume={35},
  pages={25895--25907},
  year={2022}
}

@article{verma2024induce,
  title={InduCE: Inductive counterfactual explanations for graph neural networks},
  author={Verma, Samidha and Armgaan, Burouj and Medya, Sourav and Ranu, Sayan},
  journal={Transactions on Machine Learning Research},
  year={2024}
}

@article{he2024global,
  title={Global Graph Counterfactual Explanation: A Subgraph Mapping Approach},
  author={He, Yinhan and Zheng, Wendy and Zhu, Yaochen and Ma, Jing and Mishra, Saumitra and Raman, Natraj and Liu, Ninghao and Li, Jundong},
  journal={arXiv preprint arXiv:2410.19978},
  year={2024}
}

@inproceedings{abrate2021counterfactual,
  title={Counterfactual graphs for explainable classification of brain networks},
  author={Abrate, Carlo and Bonchi, Francesco},
  booktitle={Proceedings of the 27th ACM SIGKDD conference on knowledge discovery \& data mining},
  pages={2495--2504},
  year={2021}
}

@inproceedings{numeroso2021meg,
  title={Meg: Generating molecular counterfactual explanations for deep graph networks},
  author={Numeroso, Danilo and Bacciu, Davide},
  booktitle={2021 International Joint Conference on Neural Networks (IJCNN)},
  pages={1--8},
  year={2021},
  organization={IEEE}
}

@article{wellawatte2022model,
  title={Model agnostic generation of counterfactual explanations for molecules},
  author={Wellawatte, Geemi P and Seshadri, Aditi and White, Andrew D},
  journal={Chemical science},
  volume={13},
  number={13},
  pages={3697--3705},
  year={2022},
  publisher={Royal Society of Chemistry}
}

@article{crupi2024counterfactual,
  title={Counterfactual explanations as interventions in latent space},
  author={Crupi, Riccardo and Castelnovo, Alessandro and Regoli, Daniele and San Miguel Gonzalez, Beatriz},
  journal={Data Mining and Knowledge Discovery},
  volume={38},
  number={5},
  pages={2733--2769},
  year={2024},
  publisher={Springer}
}

@article{du2025identifying,
  title={Identifying Evidence Subgraphs for Financial Risk Detection via Graph Counterfactual and Factual Reasoning},
  author={Du, Huaming and Yuan, Lei and Yang, Qing and Chen, Xingyan and Zhao, Yu and Ji, Han and Zhuang, Fuzhen and Yang, Carl and Kou, Gang},
  journal={arXiv preprint arXiv:2503.06441},
  year={2025}
}

@article{bajaj2021robust,
  title={Robust counterfactual explanations on graph neural networks},
  author={Bajaj, Mohit and Chu, Lingyang and Xue, Zi Yu and Pei, Jian and Wang, Lanjun and Lam, Peter Cho-Ho and Zhang, Yong},
  journal={Advances in neural information processing systems},
  volume={34},
  pages={5644--5655},
  year={2021}
}

@article{zannotti2026counterfactual,
  title={COunterfactual Reasoning for Temporal EXplanations: Plausible and Robust Explanations for EEG-Based Seizure Detection},
  author={Zannotti, Martina and Prenkaj, Bardh and Piangerelli, Marco and Corradini, Flavio and Kasneci, Gjergji},
  journal={Transactions on Machine Learning Research},
  year={2026}
}

@article{lavecchia2025explainable,
  title={Explainable artificial intelligence in drug discovery: bridging predictive power and mechanistic insight},
  author={Lavecchia, Antonio},
  journal={Wiley Interdisciplinary Reviews: Computational Molecular Science},
  volume={15},
  number={5},
  pages={e70049},
  year={2025},
  publisher={Wiley Online Library}
}

@article{liu2026towards,
  title={Towards faithful class-level self-explainability in graph neural networks by subgraph dependencies},
  author={Liu, Fanzhen and Ma, Xiaoxiao and Yang, Jian and Abuadbba, Alsharif and Moore, Kristen and Nepal, Surya and Paris, Cecile and Sheng, Quan Z and Wu, Jia},
  journal={IEEE Transactions on Knowledge and Data Engineering},
  year={2026},
  publisher={IEEE}
}

@article{ragno2026logic,
  title={On logic-based self-explainable graph neural networks},
  author={Ragno, Alessio and Plantevit, Marc and Robardet, C{\'e}line},
  journal={Advances in Neural Information Processing Systems},
  volume={38},
  pages={116138--116166},
  year={2026}
}

\end{document}